\documentclass[10pt,twocolumn,letterpaper]{article}

\usepackage{wacv}
\usepackage{times}
\usepackage{epsfig}
\usepackage{graphicx}
\usepackage{amsmath}
\usepackage{amssymb}

\usepackage{pgfplots}
\usepackage{tikz}
\usepackage{caption}
\usepackage{subcaption}
\usepgfplotslibrary{colorbrewer}
\pgfplotsset{cycle list/Set1-9}
\tikzset{every picture/.style={line width=1pt}}
\usetikzlibrary{patterns}
\usepackage{booktabs}
\usepackage{graphicx}
\usepackage{subfiles}
\usepackage{amsmath,amssymb} % define this before the line numbering.
\usepackage{algorithm}
\usepackage[noend]{algpseudocode}
\usepackage{bm}
\usepackage{color}
\usepackage[numbers]{natbib}
\usepackage{diagbox}
\newcommand{\C}{\mathcal{C}}
\newcommand{\Cbase}{\mathcal{C}_{\text{base}}}
\newcommand{\Cnovel}{\mathcal{C}_{\text{novel}}}
\newcommand{\SData}{\mathcal{S}}
\newcommand{\Strain}{\mathcal{S}_{\text{train}}}
\newcommand{\Stest}{\mathcal{S}_{\text{test}}}
\newcommand{\Strainnovel}{\mathcal{S}_{\text{train}}^{\text{novel}}}
\newcommand{\Sgennovel}{\mathcal{S}_{\text{gen}}^{\text{novel}}}

\wacvfinalcopy % *** Uncomment this line for the final submission

\def\wacvPaperID{632} % *** Enter the wacv Paper ID here

\usepackage{pdfpages}
\usepackage{graphicx}
\ifwacvfinal\fi
\begin{document}

%%%%%%%%% TITLE
\title{Low-Shot Learning from Imaginary 3D Model}

% Authors at the same institution
\author{Frederik Pahde\textsuperscript{1}, Mihai Puscas\textsuperscript{1,2}, Jannik Wolff\textsuperscript{1,3}\\ Tassilo Klein\textsuperscript{1}, Nicu Sebe\textsuperscript{2}, Moin Nabi\textsuperscript{1} \\
\textsuperscript{1}SAP SE., Berlin, \textsuperscript{2}University of Trento, \textsuperscript{3}TU Berlin\\
{\tt\small \{frederik.pahde, tassilo.klein, m.nabi\}@sap.com}\\
{\tt\small \{mihaimarian.puscas, nicu.sebe\}@unitn.it, jannik.wolff@campus.tu-berlin.de}
}
% Authors at different institutions
%\author{First Author \\
%\and
%Second Author \\

\maketitle
\ifwacvfinal\thispagestyle{empty}\fi

%%%%%%%%% ABSTRACT
\begin{abstract}
Since the advent of deep learning, neural networks have demonstrated remarkable results in many visual recognition tasks, constantly pushing the limits. However, the state-of-the-art approaches are largely unsuitable in scarce data regimes. To address this shortcoming, this paper proposes employing a 3D model, which is derived from training images. Such a model can then be used to hallucinate novel viewpoints and poses for the scarce samples of the few-shot learning scenario. A self-paced learning approach allows for the selection of a diverse set of high-quality images, which facilitates the training of a classifier.
The performance of the proposed approach is showcased on the fine-grained CUB-200-2011 dataset in a few-shot setting and significantly improves our baseline accuracy.

%Firstly, we propose a novel visual recognition benchmark covering the scarce data setting. Secondly, we introduce a method for one-shot fine-grained recognition using textual descriptions of visual data, which outperforms previous approaches. Therefore we propose a self-paced learning approach 

%For each visual category, we first generate a set of images by conditioning on the textual description of the category using StackGANs. Next, we rank the generated images based on their class-discriminatory power and only pick the most discriminative images to extend the dataset. Finally, a (independent) classifier can be trained using an extended training set. %In the 1-, 2-, and 5-shot scenario the discriminative hallucinated by our method improves the top-5 accuracy by 4.9 to 8.6 percentage points on the CUB dataset.
%We find that results of our method on the CUB-200-2011 dataset show improved accuracy in the 1-, 2-, and 5-shot setting when employing multimodal data.


% Since the advent of deep learning, neural networks have demonstrated remarkable results in many visual recognition tasks, constantly pushing the limits. However, the state-of-the-art approaches are largely unsuitable in scarce data regimes. To address this shortcoming, this paper proposes employing a 3D model, which is derived from training images.  This model allows for bridging the information gap by enabling the hallucination of images from the rare novel categories in the few-shot scenario. This is in conjunction with a self-paced learning approach allows for the generation of a diverse set of high-quality images facilitating the training of a discriminative classifier.
% The performance of the proposed approach is showcased on the finegrained CUB-200-2011 dataset, reaching state-of-the-art performance in the one-shot scenario. 
\end{abstract}

\section{Introduction}
% In recent years, deep learning techniques have achieved exceptional results in many domains such as computer vision and NLP. These advances can be explained by improvements to algorithms and model architecture along with increasing computational power, and in particular growing availability of big data. However, 
Since the successful introduction of deep learning techniques in countless computer vision applications, considerable research has been conducted to reduce the amount of annotated data needed for training such systems. Commonly, this data requirement problem has been approached systematically by developing algorithms which either require less expensive annotations such as semi-supervised or weakly supervised approaches, or more rigorously no annotations at all such as unsupervised systems. Although in theory quite appealing, the usual trade-off in these systems when applicable, is the overall reduced performance.\\%- in effect sidestepping the big data assumption.
%The big data assumption is key for conventional deep learning applications but often also a limiting factor. 
%However, in many applications it is too expensive (or even impossible) to acquire a sufficient number of training samples, resulting in inferior model accuracy. 
%In contrast, humans are able to
% learning is insusceptible, Furthermore, the requirement for large amounts of training data is in stark contrast to human learning, which can 
%quickly learn from only few instances. 
% This makes alternative learning approaches that require less training data an attractive research topic. 
More importantly, there exist situations where the availability of annotated data is heavily skewed, reflecting the tail distribution found in the wild. In consequence, research in the domain of low-shot learning, i.e. learning and generalizing from only few training samples, has gained more and more interest (e.g. \cite{ravi_optimization_2017,snell_prototypical_2017,vinyals_matching_2016}). As such, generative approaches for artificially increasing the training set in low-shot learning scenarios have been shown to be effective. Specifically, it was shown that with increasing quality and diversity of the generation output the overall performance of the low-shot learning system can be boosted %As such, purely visual generative methods, making use of more complex hallucination techniques 
\cite{hariharan_low-shot_2017,navaneethsemi,odena2016conditional,pahde2018discriminative}.\\%, making use of cross-modality hall where data describing the novel class is widely available, MIHCITEFred MIHRephrase have been shown to be effective. %We propose a system that has its generative capabilities maximized, reconstructing all possible viewpoints of an object from a single camera viewpoint. MIHUGH
%However, research has focused on utilizing only one data modality (mostly images) so far. 
% Overcoming this limitation and 
%By including data from additional modalities (e.g. textual descriptions) we can overcome limitations in the low data regime, resulting in improved model performance. 
%Our key assumption is that incorporating multimodal data (i.e. images and fine-grained descriptions thereof) forces the model to identify highly discriminative features \textit{across modalities} facilitating use in few-shot scenario.
% (e.g. parts in the image and textual attributes in the description).
%Specifically, pursuing multimodality suggests that novel classes with low training data in one modality can benefit from previously learned features. %This would enable them to be characterized more efficiently and accurately, resulting in improved performance in the few-shot learning setting (cf. \cite{elhoseiny_link_2017}). 
\begin{figure*}[t!]
	\centering
  \includegraphics[width=0.8\textwidth]{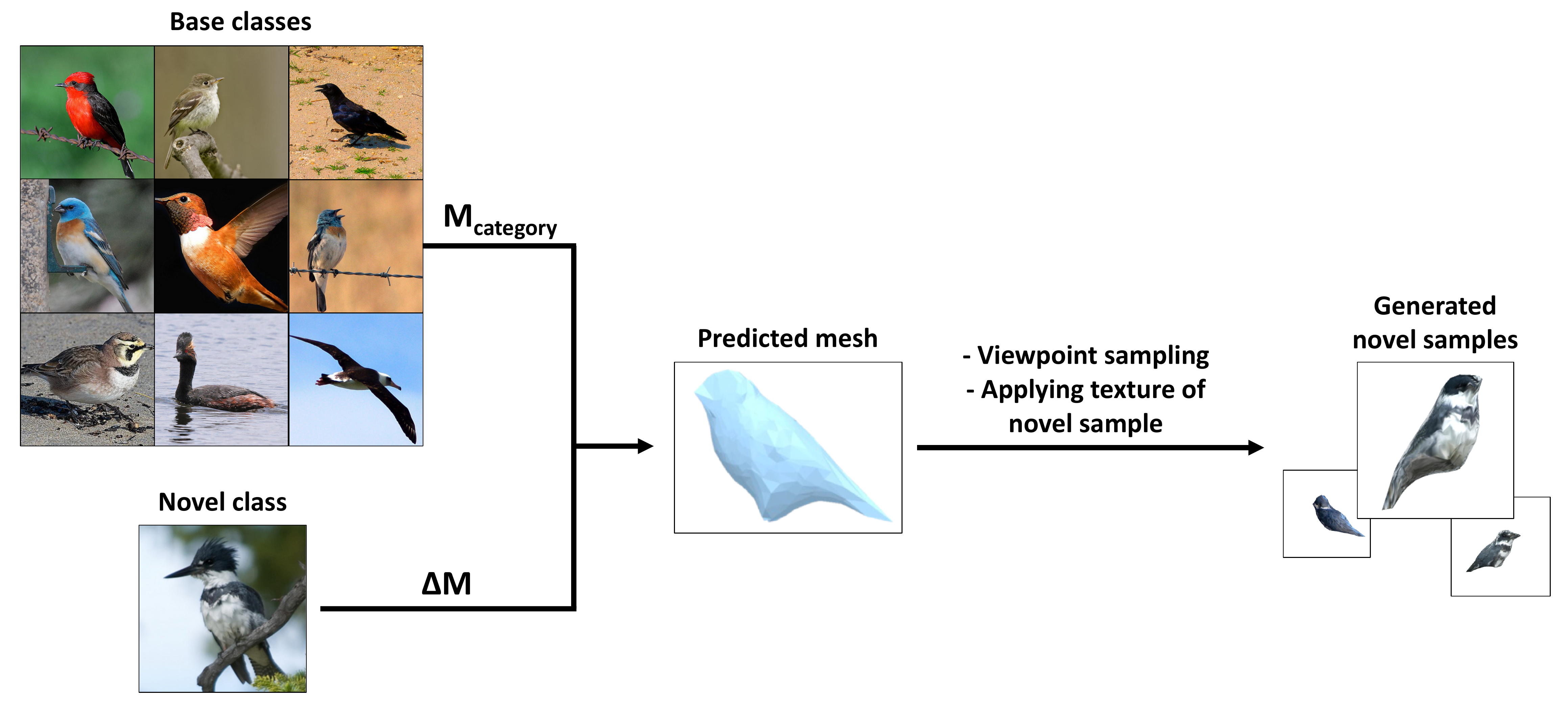}
	\caption{This figure illustrates one of our two generative methods, which is based on~\cite{kanazawa2018learning}: We first learn a generic mesh of the bird category. This mesh is then altered to fit the appearance of the target bird. We rotate the predicted 3D mesh to capture various viewpoints resulting in many 2D images that resemble the target bird. Those meshs are then coated with the novel bird's texture. To cope with the varying quality, we subsequently apply a self-paced learning mechanism, which is elaborately outlined in figure~\ref{fig:method} and in the remainder of the paper. For the second approach to sample generation, we exploit the pose variety of the base birds visible on the top left to enhance diversity. This approach is visualized in Figure~\ref{fig:lastFigure}.}
	\label{fig:1}
\end{figure*}
In this context, we propose to maximize the visual generative capabilities. 
Specifically, we assume a scenario where the base classes have a large amount of annotated data whereas the data for novel categories are scarce. To alleviate the data shortage we employ a high quality generation stage by learning a 3D structure \cite{kanazawa2018learning} of the novel class. A curriculum-based discriminative sample selection method further refines the generated data, which promotes learning more explicit visual classifiers.\\
Learning the 3D structure of the novel class facilitates low-shot learning by allowing us to hallucinate images from different viewpoints of the same object. Simultaneously, learning the novel objects' texture map allows us for a controlled transfer of the novel objects' appearance to new poses seen in the base class samples. Freely hallucinating w.r.t. different poses and viewpoints of a single novel sample then in turn allows us to guarantee novel class data diversity.
The framework by Kanazawa et al.~\cite{kanazawa2018learning} has proven to be very effective for learning both 3D models and texture maps without expensive 3D model annotations.
While reconstructing a 3D model from single images in a given category has been achieved in the past \cite{vicente2014reconstructing, kar2015category}, these methods lack easy applicability to a hallucinatory setup and specifically miss any kind of texture and appearance reconstruction. The intuition behind our idea is visualized in Fig.~\ref{fig:1}\\
With a broad range of images generated for varying viewpoints and poses for the novel class, a selection algorithm is applied. 
To this end, we follow the notion of \emph{self-paced learning} strategy, which is a general concept that has been applied in many other studies \cite{kumar2010self,sangineto2016self}. It is related to curriculum learning \cite{bengio2009curriculum}, and is biologically inspired by the common human process of gradual learning, starting with the simplest concepts and increasing complexity.
We employ this strategy to select a subset of images generated from the imaginary 3D model, which are associated with high confidence w.r.t. ``class discriminativeness'' by the discriminator. 
%This discriminator is pre-trained on all available data from base classes, for which we have many available samples. Fine-tuning is then performed using data from the novel classes, which live in a scarce data regime, and the subset of generated images of sufficient quality. One part of the hallucinatory process of generating new samples is portrayed in Figure~\ref{fig:1}, the other one in Figure~\ref{fig:lastFigure}.
%This subset of generated images is progressively increased in subsequent iterations when the model becomes more mature and is able to capture more complexity. The resulting self-paced process allows to handle uncertainty related to the quality of generated samples.
%In this paper the strategy is employed to select a subset of images generated from the imaginary 3D model, which are associated with high confidence w.r.t. ``class discriminativeness'' by the discriminator. 
 Specifically the self-pacedness allows to handle the uncertainty related to the quality of generated samples. Here the notion of ``easy'' is interpreted as ``high quality''. Training is then performed using only the subset consisting of images of sufficient quality. This set is then in turn progressively increased in the subsequent iterations when the model becomes more mature and is able to capture more complexity.\\
The main contributions of this work are: 
\textbf{First}, we massively expand the diversity of generating data from sparse samples of novel classes through learning 3D structure and texture maps.
\textbf{Second}, we leverage a self-paced learning strategy facilitating reliable sample selection.\\
Our approach features robustness and outperforms the baseline in the challenging low-shot scenario.

\begin{figure*}[t]
	\centering
  \includegraphics[width=0.8\textwidth]{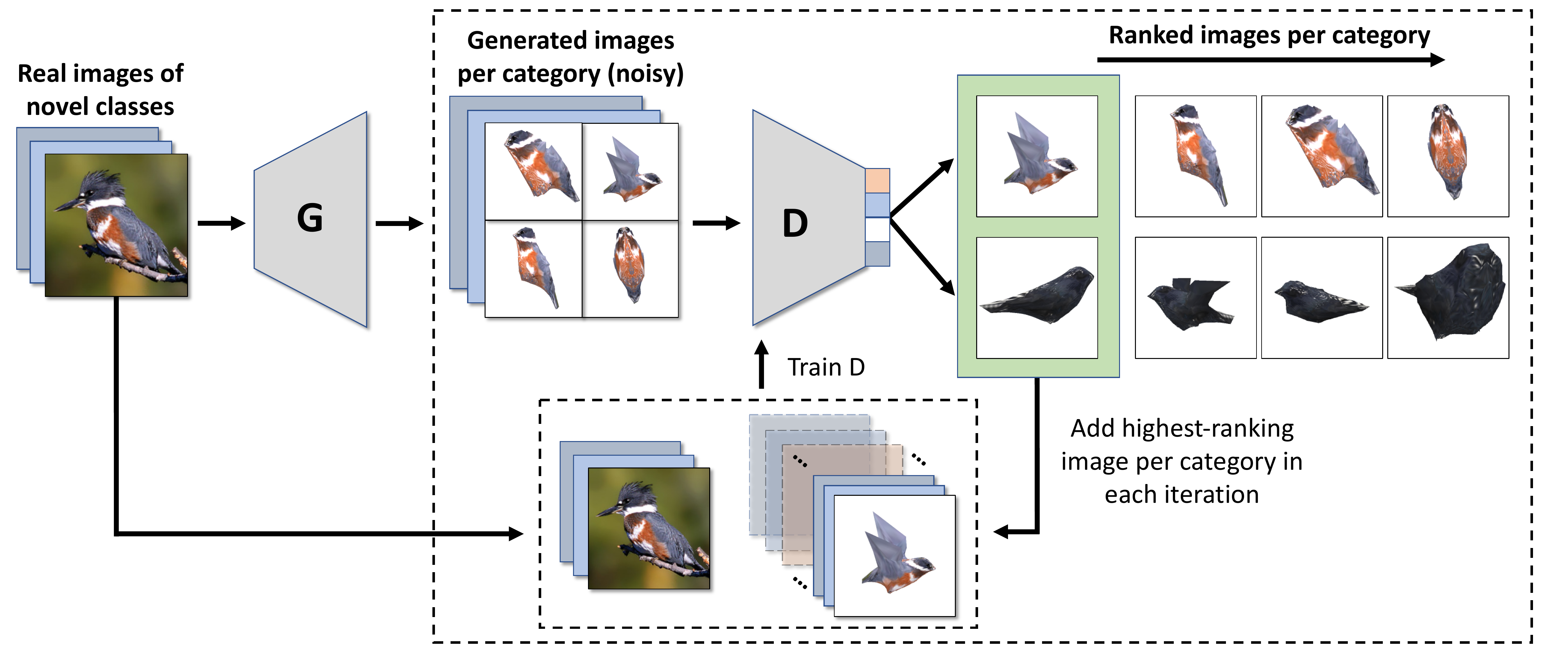}
	\caption{Self-paced fine-tuning on novel classes: For each novel class, noisy samples are generated with different viewpoints and poses by $G$. Those images are ranked by $D$ based on their class-discriminatory power. The highest-ranking images are added to the novel samples and used to update D, which is trained using a simple cross-entropy loss. This process is repeated multiple times. Initially, $D$ has been pre-trained on all base class data.}
	\label{fig:method}
\end{figure*}
\section{Related Work}
In this section we briefly review previous work considering: (1) low-shot learning, (2) 3D model learning and inference and (3) self-paced learning.

\subsection{Low-Shot Learning}
For learning deep networks using limited amounts of data, different approaches have been developed. Following Taigman et al.~\cite{taigman2014deepface}, Koch et al.~\cite{koch_siamese_2015} interpreted this task as a verification problem, i.e. given two samples, it has to be verified, whether both samples belong to the same class. Therefore, they employed siamese neural networks \cite{bromley1994signature} to compute the distance between the two samples and perform nearest neighbor classification in the learned embedding space. Some recent works approach few-shot learning by striving to avoid overfitting by modifications to the loss function or the regularization term. Yoo et al.~\cite{yoo_efficient_2017} proposed a clustering of neurons on each layer of the network and calculated a single gradient for all members of a cluster during the training to prevent overfitting. The optimal number of clusters per layer is determined by a reinforcement learning algorithm. A more intuitive strategy is to approach few-shot learning on data-level, meaning that the performance of the model can be improved by collecting additional related data. Douze et al.~\cite{douze_low-shot_2017} proposed a semi-supervised approach in which a large unlabeled dataset containing similar images was included in addition to the original training set. This large collection of images was exploited to support label propagation in the few-shot learning scenario. Hariharan et al.~\cite{hariharan_low-shot_2017} combined both strategies (data-level and algorithm-level) by defining the squared gradient magnitude loss, that forces models to generalize well from only a few samples, on the one hand and generating new images by hallucinating features on the other hand. For the latter, they trained a model to find common transformations between existing images that can be applied to new images to generate new training data (see also \cite{wang_low-shot_2018}). Other recent approaches to few-shot learning have leveraged meta-learning strategies. Ravi et al.~\cite{ravi_optimization_2017} trained a long short-term memory (LSTM) network as meta-learner that learns the exact optimization algorithm to train a learner neural network that performs the classification in a few-shot learning setting. This method was proposed due to the observation that the update function of standard optimization algorithms like SGD is similar to the update of the cell state of a LSTM. Bertinetto et al.~\cite{bertinetto_learning_2016} trained a meta-learner feed-forward neural network that predicts the parameters of another, discriminative feed-forward neural network in a few-shot learning scenario. Another tool that has been applied successfully to few-shot learning recently is attention. Vinyals et al.~\cite{vinyals_matching_2016} introduced matching networks for one-shot learning tasks. This network is able to apply an attention mechanism over embeddings of labeled samples in order to classify unlabeled samples. One further outcome of this work is that it is helpful to mimic the one-shot learning setting already during training by defining mini-batches, called episodes with subsampled classes. Snell et al.~\cite{snell_prototypical_2017} generalize this approach by proposing prototypical networks. Prototypical networks search for a non-linear embedding space (the prototype) in which classes can be represented as the mean of all corresponding samples. Classification is then performed by finding the closest prototype in the embedding space. In the one-shot scenario, prototypical networks and matching networks are equivalent. 

\subsection{3D Shape Learning}

Inferring the 3D shape of an object from differing viewpoints has long been a topic of interest in computer vision.
Based on the idea that there exists a categorical-specific canonical shape, and that class-specific deformations of it can be learned, systems such as SMPL~\cite{loper2015smpl} and "Keep it SMPL"~\cite{bogo2016keep} model a human 3D shape space, while Zuffi et al.~\cite{zuffi20173d} perform a similar task for quadruped animals.
However, even though these methods are able to use synthetic training data, they still rely on a 3D shape ground truth.
In contrast, Kanazawa et al.~\cite{kanazawa2018learning} make use of much cheaper keypoint and segmentation mask annotations, which allows both 3D mesh and texture inference for images.

%Making use of much cheaper keypoint and segmentation mask annotations, \cite{kanazawa2018learning} allows both 3D Mesh and texture inference for sample images. 

\subsection{Self-Paced Learning}
Recently, many studies have shown the benefits of organizing the training examples in a meaningful order (e.g., from simple to complex) for model training. Bengio et al. \cite{bengio2009curriculum} first proposed a general learning strategy: curriculum learning. They show that suitably sorting the training samples, from the easiest to the most difficult, and iteratively training a classifier starting with a subset of easy samples (which is progressively augmented with more and more difficult samples), can be useful to find better local minima. %In \cite{chen2015webly}, easy and difficult images (taken from datasets known to be more or less “difficult”) are provided for training a CNN in order to learn generic CNN features using webly annotated data. 
Note that in this and in all the other curriculum-learning-based approaches, the order of the samples is provided by an external supervisory signal, taking into account human domain-specific expertise.\\
Curriculum learning was extended to self-paced learning by Kumar et al. \cite{kumar2010self}. They proposed the respective framework, automatically expanding the training pool in an easy-to-hard manner by converting the curriculum mechanism into a concise regularization term. Curriculum learning uses human design to organize the examples, and self-paced learning can automatically choose training examples according to the loss. Supancic et al. \cite{supancic2013self} adopt a similar framework in a tracking scenario and train a detector using a subset of video frames, showing that this selection is important to avoid drifting.
Jiang et al.~\cite{jiang2014self} pre-cluster the training data in order to balance the selection of the easiest samples with a sufficient inter-cluster diversity. Pentina et al.~\cite{pentina2015curriculum} propose a method in which a set of learning tasks is automatically sorted in order to allow a gradual sharing of information among tasks. In Zhang et al.'s~\cite{zhang2017bridging} model saliency is used to progressively select samples in weakly supervised object detection.
In context of visual categorization some of these self-paced learning methods use CNN-based features to represent samples~\cite{liang2015towards} or use a CNN as the classifier directly~\cite{sangineto2016self}.  
%In contrast, we additionally formulate the GAN training in the self-paced learning protocol allowing to optimize the generator to create samples with the amount of class-discriminative features necessary at a certain step of the training.

%focus on a generative network that produces high-quality and diverse samples. This heterogeneity comes at the expense of noisy samples, which in turn can provide satisfactory diverse samples in order to learn a powerful classifier in few-shot scenarios.
%which in turn create a setting in which self-paced strategies can flourish.

%Although some of these self-paced methods use pre-trained CNN-based features to represent samples (e.g., \cite{liang2015towards}), or uses a CNN as the classifier directly(e.g., \cite{sangineto2016self}), none of them formulates the self-paced strategy in a GAN training protocol as we do in this paper.

%\subsection{Self-Paced Learning}
%\section{Background on GANs}
%\subfile{sections/background}
\section{Method}
\subsection{Preliminaries}
In this subsection we introduce the necessary notation.\\
Let $\mathcal{I}$ denote the image space, $\mathcal{T}$ the texture space , $\mathcal{M}$ the 3D mesh space and $\C=\lbrace 1, ..., L\rbrace$ the discrete label space. Further, let $x_i \in \mathcal{I}$ be the $i$-th input data point, and $y_i \in \C$ its label.
In the low-shot setting, we consider two subsets of the label space: $\Cbase$ for labels for which we have access to a large number of samples, and the novel classes $\Cnovel$, which are underrepresented in the data. Note that both subsets exhaust the label space $\C$, i.e. $\C = \Cbase \cup \Cnovel$. We further assume that in general $|\Cnovel| \ll |\Cbase|$.\\
The dataset $\SData$ decomposes as follows: $\SData = \Strain \cup \Stest$, $\Strain \cap \Stest = \emptyset$.
The training data $\Strain$ consists of 2-tuples $\{(x_i, y_i)\}_{i=1}^{N}$ taken from the whole data set containing both image samples and labels. Furthermore, for 3D model prediction we also attach 3-tuples $\{(l_i, k_i, m_i)\}_{i=1}^{N}$, with $l_i$ being a foreground object segmentation mask and $k_i$ a 15-point keypoint vector representing the pose of the object. Additionally, $m_i$ denotes the weak-perspective camera, which is estimated by leveraging structure-from-motion on the training instances' keypoints $k_i$. The test data is drawn from the novel classes and does not contain any 3D information, but solely images and their labels.
Next, there is also $\Strainnovel = \{(x_i, y_i, l_i, k_i, m_i) : (x_i, y_i, l_i, k_i, m_i) \in \Strain, y_i \in \Cnovel\}_{i=1}^M \subset \Strain$, which denotes the training data for the novel categories.
For each class in $\Cnovel$, $k$ samples can be used for training (k-shot), resulting in $\left|\Strainnovel\right|\ll\left|\Strain\right|$

% The test data consists for tuples $\Stest = \{(x_i, y_i) : y_i \in \Cnovel\}_{i=1}^m$, which belong to novel classes. 

%Consequently, in accordance with the low-shot scenario $k = \left|\Strainnovel\right|\ll\left|\Strain\right| = n$. Additionally, in a low-shot learning scenario, the number of samples per category of $\Cbase$ may be further limited to be $g$, which is denoted by $\Strainnovel(g)$.

\begin{algorithm}
	\caption{Self-paced learning, \textsc{RANK}() is a function that ranks generated images based on their score of $D'$ and \textsc{TOP}() returns the highest ranked images}\label{spl}
\begin{algorithmic}[1]
	\State \textbf{Input:} Pre-trained network $D$, $S^{novel}_{gen}$, $r$
	\State \textbf{Output:} Fine-tuned classifier $D'$
	\For{$i = 1,\ldots, n$}
	\State $\mathcal{S}_{\text{all}}^{\text{novel}} = \emptyset$
	\For{$c \in \Cnovel$}
    	\State $\text{candidates}=\emptyset$
        \For{$x_i^{gen} \in S^{novel}_{gen}$}
        	\State $\text{candidates} = \text{candidates}\cup x_i^{gen}$
        \EndFor
% 		\text{} \in S^{novel}_{gen}
		\State $\text{candidates}_{\text{ranked}}$ $=$ \textsc{rank}$(\text{candidates}, D')$
		\State $\text{sample} = \textsc{top}(\text{candidates}_{\text{ranked}}, r)$
		\State $\Sgennovel = \Sgennovel \cup \text{sample}$
	\EndFor
	\State $\mathcal{S}_{\text{all}}^{\text{novel}} = \Strainnovel \cup \Sgennovel$
	\State update $D'$ with $\mathcal{S}_{\text{all}}^{\text{novel}}$
	\EndFor
\end{algorithmic}
\end{algorithm}

\subsection{3D Model Based Data Generation}
\label{subsec:3d}

The underlying observation on which our method is based on is that increased diversity of generated images directly translates into higher classification performance for novel categories.
% this diversity can be guaranteed by learning a 3d model...
The proposed work aims at emulating processes in human cognition that allow for reconstructing different viewpoints and poses through conceptualizing a 3D model of an object of interest. Specifically, we aim to learn such a 3D representation for novel samples appearing during training and leverage it to predict different viewpoints and poses of that object.

% Given a single image $x_i$, we predict $f_{mesh}(x_i) = (M, T)$, where $M$ is a mesh that captures the 3D structure of the object represented, and $T$ is a predicted texture map for the given object.
%Because of this, predicting the texture of an instance $x_i$ can be formulated as predicting the texture of the mean shape $M_{cat
%}$.
We use the architecture proposed by Kanazawa et al.~\cite{kanazawa2018learning} to predict a 3D mesh $M_i$ and texture $T_i$ from an image sample $x_i$.
With the assumption that all $x_i \in \mathcal{I}$ represent objects of the same category, the shape of each instance is predicted by deforming a learned category-specific mesh $M_{cat}$. Note that \textit{category} refers to the entire fine-grained bird dataset, as opposed to \textit{class}. All recovered shapes will share a common underlying 3D mesh structure, $M_{i} = M_{cat} + \Delta M_{i}$, with $\Delta M_{i}$ being the predicted mesh deformation for instance $x_i$. Because the mesh $M$ has the same vertex connectivity as the average categorical mesh $M_{cat}$, and further as $M_{sphere}$ representing a sphere, a predicted texture map $T_i$ can be easily applied over any generated mesh.

%Using Kanazawa et al.'s model, learning 3D representations does not require expensive 3D model or multi-view annotations, which constitutes an advantage over related methods. The system minimizes the distances between a rendered object mask, keypoints and texture to their respective ground truth, using only images representing different instances of a category. \\
An advantage of \cite{kanazawa2018learning} over related methods is that learning the 3D representation does not require expensive 3D model or multi-view annotations.\\ 
Given $(M_i, T_i, \Theta_i)$ and $\Theta = (\alpha, \beta, \gamma$), where the three camera rotation angles $\alpha, \beta, \gamma$ are sampled uniformly from $\left[0, \pi/6\right]$, we can project the reconstructed object using $f_{gen}(M_i, T_i, \Theta_i)$ such that $X^{view}_i = \{x^0_i, ..., x^L_i\}$ contains samples of the object seen from different viewpoints.

As $X^{view}_i$ only contains different viewpoints of the novel object, it will not contain any novel poses. This is a concern for non-rigid object categories, where it cannot be guaranteed that the unseen samples in a novel class will have similar poses to the known samples in the novel class. To mitigate this, the diversity of the generated data must be expanded to include new object poses.

All meshes predicted from $x_j \in S_{base}$ obtain the spherical texture map $T_i$ corresponding to $x_i \in \Strainnovel$ using $f_{gen}(M_j, T_i, \Theta_j)$. This transfers the shape from base class objects to novel class instances resulting in $X^{pose}_i = \{x^j_i, ..., x^S_i\}$.

\begin{figure}[ht]
	\centering
  \includegraphics[width=1\columnwidth]{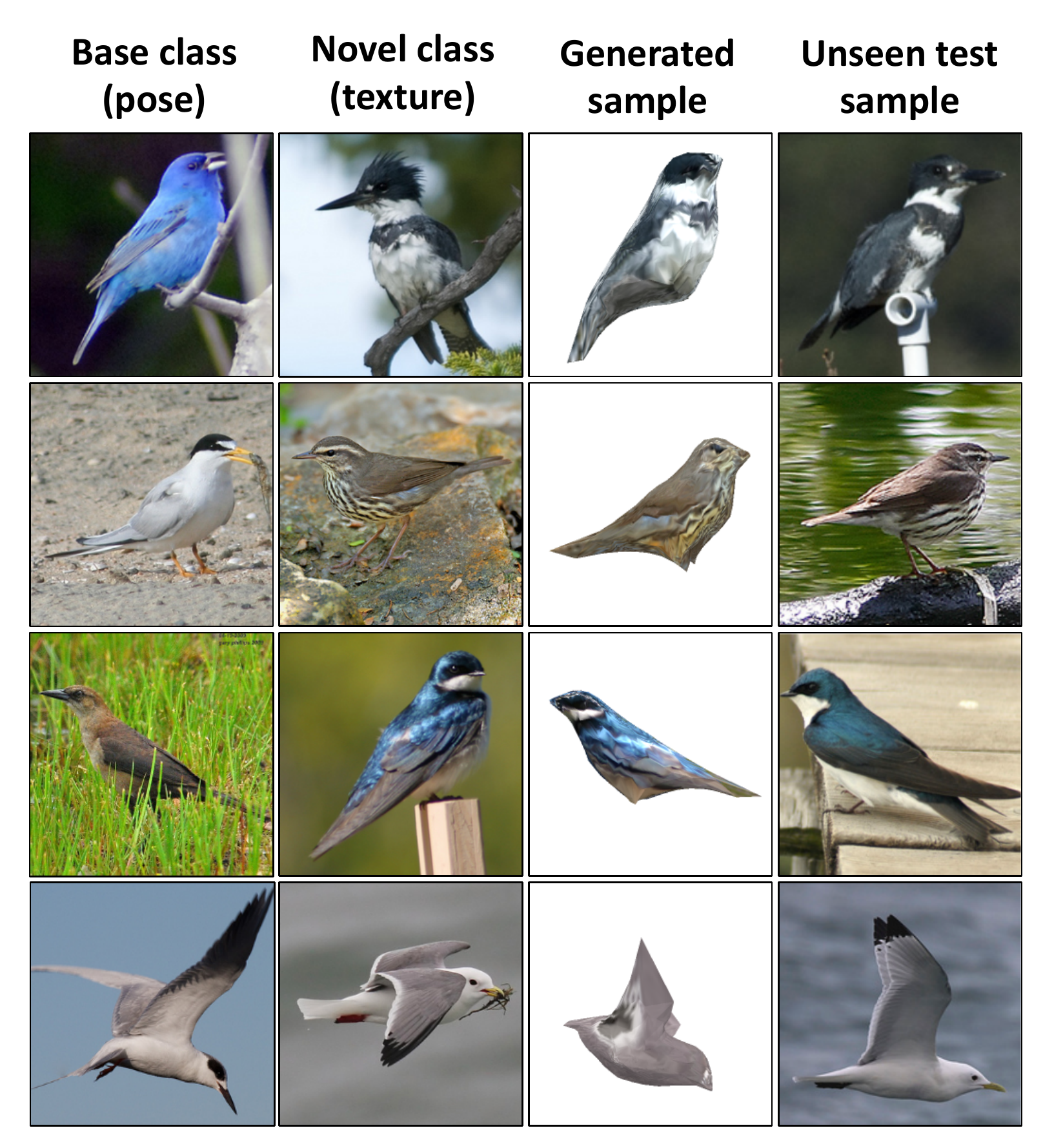}
	\caption{
	Texture from novel class birds is transferred onto poses from base class birds. The generated samples have been previously selected by the discriminator w.r.t. to their class-discriminatory power in the self-paced learning setting. Those hallucinations are visually similar to unseen test samples, indicating their value for training a classifier.}
	\label{fig:lastFigure}
\end{figure}

%We apply the spherical texture map $T_i$, corresponding to $x_i \in \Strainnovel$ to all meshes predicted from $S_{base}$, transferring the appearance of an object in a novel class to objects in the base classes. 
%Within the fine-grained context, we assume that the pose space/kinematics of any sample in the dataset will coincide.
Using poses from images of different labels is an inherently noisy approach through inter-class mesh variance. However, a subsequent sample selection strategy allows the algorithm to make use of the most representative poses. Indeed, as seen in Figure~\ref{fig:lastFigure}, meshes $M_j \in S_{base}$ exist for which the predicted images $x^j_i$ are visually similar to samples of the unseen classes.

Thus, for each sample $x_i \in S_{train}^{novel}$, a set of images $S_{gen}^{novel} = X^{view}_i \cup X^{pose}_{i}$ is generated. This generated data captures both different viewpoints of the novel class and the appearance of the novel class applied to differing poses from the base classes.

\subsection{Pre-Training of Classifier}
\label{subsec:auxclassifier}

In the low-shot learning framework proposed by Hariharan and Girshick~\cite{Girshick_lowshot}, a representation of the base categorical data must be learned beforehand. This is achieved by learning a classifier on the samples available in the base classes, i.e. $x_i \in S_{train}^{base}$. For this task we make use of an architecture identical to the StackGAN discriminator \cite{zhang_stackgan++:_2017}, modified to serve as a classifier. This discriminator $D$ is learned on $S_{train}^{base}$ by minimizing $L_{class}$ defined as a cross-entropy loss.

However, to accommodate for the different amount of classes in base and novel, D has to be adapted. Specifically, the class-aware layer with $|\Cbase|$ output neurons is replaced and reduced to $|\Cnovel|$ output neurons, which are randomly initialized. 
We refer to this adapted classifier as $D'$. 
Subsequently, the network can be fine-tuned using the available novel class data.
% Optimization is performed on the loss defined as, 

% \begin{eqnarray}
% \text{subject to:} &  0 \leq \alpha_X \leq 1, \quad \sum_{X \in {X}_{\text{novel}}} \alpha_I \leq K,
% \label{eq:2}
% \end{eqnarray}
% where $\mathcal{L}$ is the classification loss on the joint set of generated and training data $\Strainnovel \cup \Sgennovel$,
% % from $\Cnovel$. 
% ,and $\alpha_T$ is a soft selector for the images $X_{novel}$ generated from the predicrted 3d mesh and texture map $(M_i, T_i)$
% Since each $\alpha_X$ is a selector for the generated data, $K$ specifies 
% % the budget 
% the maximum number of generated
% of samples to be included in $\Sgennovel$ for the next finetuning step. Our pseudo-code is given in algorithm~\ref{spl}.

\subsection{Self-Paced Learning}
\label{subsec:spl}

As seen in section \ref{subsec:3d}, for a given novel sample $x_i \in \Strainnovel$ we can generate $S_{gen}^{novel} = X^{view}_i \cup X^{pose}_{i}$, containing new viewpoints and poses of the given object.

For the self-paced learning stage, we fine-tune with the novel samples, as well as the samples generated through projecting the predicted 3D mesh and texture maps. i.e. with the data given by $S_{train}^{novel} \cup S_{gen}^{novel}$. 

Unfortunately, the samples contained in $S_{gen}^{novel}$ can be noisy for a variety of reasons: failure in predicting the 3D mesh deformation due to a too large difference between the categorical mesh and the object mesh, or even viewpoints that are not representative to the novel class.

To mitigate this we propose a self-paced learning strategy ensuring that only the best generated samples within $\Sgennovel$ are used. 

% To obtain a meaningful ranking in the self-paced learning phase, $D'$ has to be initialized on novel classes. 
Again taking into account the setting of low-shot learning, we restrict the number of samples per class available to $k$.
% Therefore, $\Strainnovel(n)$ containing only $n$ samples per class are used to pre-train $D'$ in accordance with a low-shot learning scenario. 
Due to the limited amount of samples, the initialized $D'$ will be weak on the classification task, but sufficiently powerful for performing an initial ranking of the generated images. For this task we employ the softmax activation for class-specific confidence scoring. As $D'$ learns to generalize better, more difficult samples will be selected. 

This entails iteratively choosing generated images that have highest probability in $D'$ for $\Cnovel$, yielding a curated set of generated samples $\Sgennovel$. An issue in selecting the highest scoring sample in each iteration is the possibility of not making full use of the available data w.r.t. its diversity - the highest scoring images being of a very similar pose and viewpoint to the original sample. 

We address this shortcoming by a using a clustering-and-discard strategy: For the novel class training sample $x_i$, we generate $X^{gen}_i = \{x^0_i, ... x^{L+S}_i\}$ new images, representing new viewpoints and poses of the object.  $X^{gen}_i$ is then further associated with  $K^{gen}_i = \{k^0_i, ... k^Q_i\}$, representing all the predicted keypoints of the associated generated samples. $K^{gen}_i$ is clustered using a simple k-means implementation \cite{scikit-learn}.
On every self-paced iteration, the pose cluster associated to the selected top-ranked sample is discarded to increase data diversity.

Finally, we aggregate original samples and generated images $\Strainnovel \cup \Sgennovel$ for training, during which we update $D'$. 
% Specifically, backpropagation comprises updating $D'$ and $G$, respectively. 
Doing so yields both a more accurate ranking as well as higher class prediction accuracy as the number of samples increases.
% Consequently, iterative improvements are two-fold. 
% On the one hand, the ranking gets more accurate $D'$. 
% On the other hand, the class-discriminativeness of the samples increases. 
Ultimately, the approach learns a reliable classifier that performs well in low-shot learning scenarios. It is summarized in algorithm~\ref{spl}.

%\subsubsection{Self-paced Adversarial Training}
%The ability to rank generated images with the pre-trained $D'$ allows for data selection and guided optimization. 
%In our approach we specifically follow a self-paced learning strategy. 
%This entails iteratively choosing generated images that have highest probability in $D'$ for $\Cnovel$, yielding a curated set of high-quality generated samples $\Sgennovel$. 
%Finally, we aggregate original samples and generated images $\Strainnovel \cup \Sgennovel$ for training, during which we alternately update $D'$ and $G$. 
% Specifically, backpropagation comprises updating $D'$ and $G$, respectively. 
%Doing so yields both a more accurate ranking as well as higher class prediction accuracy as the number of samples increases.
% Consequently, iterative improvements are two-fold. 
% On the one hand, the ranking gets more accurate $D'$. 
% On the other hand, the class-discriminativeness of the samples increases. 
%Ultimately, the approach summarized in algorithm~\ref{spl} learns a reliable classifier that performs well in low-shot learning scenarios.
% Taken together, this facilitates learning a reliable classifier in low-shot learning scenario. The proposed approach is summarized in algorithm~\ref{spl}.  

\section{Experiments}
\begin{table*}
    \setlength{\tabcolsep}{5pt}
	\centering{
		\begin{tabular}{lccccc} \toprule
			& & & \textbf{k}	& 	&\\
			\textbf{Model}  & \textbf{1} & \textbf{2} & \textbf{5} & \textbf{10} & \textbf{20}\\ \midrule
			Baseline  	& 27.55&	30.75&	54.25&	58.51&	71.62	\\
            %Baseline + knn  	& 40.79	& 39.47 & 56.07	& 66.25	& 73.08	\\
            Views + poses & 33.40&	43.72&	54.81&	65.27&	74.06 \\
            SPL w/ views & 33.54&	41.49&	54.88&	65.48&	\textbf{74.97}\\
            SPL w/ poses  	& 33.82&	42.47&	54.95&	64.85&	73.64 \\
            SPL w/ poses + clustering  	& 33.40&	45.05&	57.74&	65.69&	74.62 \\
            SPL w/ poses + views    	& 35.29&	41.98&	55.37&	66.04&	71.48\\
            SPL w/ poses + views (balanced)  	& 35.77&	44.56&	54.60&	64.30&	74.83 \\
            SPL w/ all 	& \textbf{36.96}&	\textbf{45.40}&	\textbf{58.09} &	\textbf{66.53} &	74.83
			 \\\bottomrule
		\end{tabular}
	}
	\captionsetup{position=above}
	\captionof{table}{Ablation study of our model in a top-5, 50-way scenario on the CUB-200-2011 dataset in different k-shot settings, best results are in bolt.
    We observe that each of the proposed extensions increases the accuracy in
    at least one setting which justifies their usage. This regards to both, methods for generating additional data and the approach to only select generated samples of sufficient quality for training the classifier.}
	\label{tab:results}
\end{table*}

\subsection{Datasets}
We test the applicability of our method on CUB-200-2011 \cite{WahCUB_200_2011}, a fine-grained classification datasets containing 11,788 images of 200 different bird species of size $\mathcal{I} \subset \mathbb{R}^{256\times256}$. 
The data is split equally into training and test data. 
As a consequence, samples are roughly equally distributed, with training and test each containing $\approx 30$ images per class. 
Additionally, foreground masks, semantic keypoints and angle predictions are provided by \cite{kanazawa2018learning}. Note that nearly 300 images are removed where the number of visible keypoints is less or equal than 6.

Following Zhang et al.~\cite{zhang_stackgan++:_2017}, we split the data such that $\left|C_{base}\right|=150$ and $\left|C_{novel}\right|=50$. 
%Furthermore, data of the novel classes equally in training and test data. 
To simulate low-shot learning, $k\in\{1,2,5,10,20\}$ images of $C_{novel}$ are used for training, as proposed by \cite{hariharan_low-shot_2017}.

\subsection{Algorithmic Details}
During representation learning, we train an initial classifier on the base classes for $600$ epochs and use Adam \cite{kingma2014adam} for optimization.
We set the learning rate $\tau$ to $10^{-3}$ and the batch size for $D$ to 32. 
In the initialization phase for self-paced learning, we construct $D'$ by replacing the last layer of $D$ by a linear softmax layer of size $\left|C_{novel}\right|$. 
The resulting network is then optimized using the cross-entropy loss function and an Adam optimizer with the same parameters. 
Batch size is set to 32 and training proceeds for 20 epochs. 
Self-paced learning of $D'$ continues to use the same settings, i.e. the Adam optimizer minimizing a cross-entropy loss. 
In every iteration we choose exactly one generated image per class and perform training for 10 epochs.

\subsection{Models}
In order to asses the performance of individual components, we perform an ablation study.

The simplest transfer learning approach is making use of a pre-trained representation and then fine-tuning that model on the novel data. 
A first baseline (\textbf{Baseline}) uses this strategy:  we pre-train a classifier $D$ on the base classes, following by fine-tuning with $k$ novel class instances $x_i \in \Strainnovel$. This strategy makes use of the fine-grained character of the dataset, learning initial representations on $\Cbase$ and performing classification on $\Cnovel$.

A second model \textbf{views + poses} studies the validity of the generated viewpoint and pose data. For $r$ sampling iterations, a single uniformly sampled $x_i \in S^{novel}_{gen}$ is attached to a novel sample set.
%$S_{train}^{novel}(n)$, with $n$ representing the number of samples selected for each novel class. 

We then introduce sample selection to our method. Note that viewpoint generation is achieved through 3D Mesh $M_i$ and texture $T_i$ of the same sample $x_i$, while the different poses are generated through applying the novel class instance texture $T_i$ to base class meshes $M_j$. The \textbf{SPL w/ views} and \textbf{SPL w/ poses} sample the generated data from the generated viewpoints $X^{view}$ and $X^{pose}$ respectively.

\textbf{SPL w/ poses + views} makes use of the entirety of $S_{gen}^{novel}$, while \textbf{SPL w/ poses + views (balanced)} tackles the data imbalance between different viewpoint samples and different pose samples by ranking the two branches separately, and selecting one sample from each such that for one novel sample, $x_i^{max, pose}$ and $x_i^{max, view}$ are used in fine-tuning.

The clustering-and-dismissal mechanism detailed in \ref{subsec:spl} is evaluated in the \textbf{SPL w/ poses + clustering} model, while \textbf{SPL w/ all} makes use of the method in its entirety. 

%We apply this strategy on a first baseline (\textbf{Finetuning}), for which we pre-train a classifier $T$ that has exactly the same architecture as $D$ on the base classes, followed by finetuning with the few instances of novel classes on $\Strainnovel$. 
% This meta-learning strategy learns meaningful representations on the base classes $\Cbase$ that can be used for novel classes $\Cnovel$. 
% A second baseline (\textbf{Initialization}) constitutes our first contribution. 
% We modify the discriminator $D$ of the StackGAN, which we obtain from the representation learning phase, to obtain $D'$ by exchanging the discriminator's last layer. 
% Finetuning is then performed on the real samples from novel classes $\Strainnovel$. 
% Note that the \textit{initialization} baseline uses $D$ which is pre-trained using the adversarial principle during the StackGAN training, in contrast to the \textit{finetuning} baseline that uses $T$ as pre-trained by a conventional classifier.  
%It should be noted in contrast to the \textit{finetuning}, for the fourth baseline the initialized $D'$ was pre-trained in an adversarial fashion during the StackGAN training. 
% Afterwards, we iteratively add high-quality generated samples for  novel categories $\Sgennovel$ as described. 
% In a first self-paced experiment (\textbf{SPL-D'}) we update $D'$ using selected generated samples in every iteration. 
% In a second experiment, additionally to updating $D'$, we update $G$ in every iteration in order to be fully self-paced (\textbf{SPL-D'G}).

\subsection{Results of Ablation Study}
The results of the ablation study outlined in the previous section are shown in Table~\ref{tab:results}, presenting 50-way, top-5 accuracies for k-shot learning with $k \in \{1, 2, 5, 10, 20\}$.

We first evaluate the baseline model, which is trained on the base classes and fine-tuned on the novel classes. Due to using a relatively shallow classification network, and the sparsity of the novel samples, the network rapidly overfits.

Introducing more data diversity to the fine-tuning stage through 3D model inference provides a significant boost in performance in all $k \in \{1, 2, 5, 10, 20\}$
With the generated samples selected randomly, the network does not easily overfit, but this selection method provides no protection against noisy generated samples.

Subsequent models evaluate different selection strategies across the two defined generated data splits for new viewpoints and poses, i.e. $X^{view}$ and $X^{pose}$. 
The contribution of the self paced learning strategy can be evaluated directly comparing the top-5 accuracies of the \textbf{view + poses} model and the \textbf{SPL w/ views + poses} model. The increase of performance when k is small shows that the selection strategy can achieve better performance, but inconsistently across different $k$ values. 

One cause of this problem is how the generated data is split, and whether the classifier has access to the most valuable generated samples.  In \textbf{SPL w/ poses} and \textbf{SPL w/ views}, we only select samples from $X^{pose}$ and $X^{view}$ respectively. The experimental results of both models are similar and inferior to \textbf{SPL w/ views + poses}, where both sets are used. Even with higher performance, the aggregate model selects from $X^{view}$ almost exclusively, hinting on a type of mode collapse.

To further diversify the possible data picks, we "balance" the two sets: For each sample, $x_i^{max, pose}$ and $x_i^{max, view}$ are selected as the highest scoring samples in their respective sets. This disentangling of pose and viewpoint data offers an across-the-board improvement, as seen in \textbf{SPL w/ views + poses (balanced)}.

While normally each sample that was selected in self-paced iteration $r$ is discarded, this will likely leave a number of samples that are similar in pose, such that the classifier may rank them as maximum. This does not add significant new information to the learning process, and as such the clustering-by-pose method guiding the sample dismissal is introduced. Indeed, as observed in \textbf{SPL w/ all}, both the sample-discard strategy, and the balancing strategy are similar useful for selections in self-paced learning. With all discussed techniques introduced, the model achieves a significant performance boost compared to the baseline.

\subsection{Analysis of Self-Paced Fine-Tuning}

\begin{table}
    \setlength{\tabcolsep}{5pt}
	\centering{
		\begin{tabular}{lccccc} \toprule  
                  & Baseline & NN  & Our (shallow) & Our (ResNet) & \cite{hariharan_low-shot_2017} \\ \midrule
               & 9.1 & 9.7 & 14.4 & 18.5 & 19.1 \\
			 \bottomrule
		\end{tabular}
	}
	\captionsetup{position=above}
	\captionof{table}{Top-1, 50-way, 1-shot accuracies on the CUB-200-2011 dataset. We see that our shallow CNN (trained with self-paced learning) exceeds both baselines. The ResNet (not trained with self-paced learning) is within reach of Hariharan and Girshick's model with SGM loss~\cite{hariharan_low-shot_2017}, for which we have reproduced respective results.}
	\label{tab:results2}
\end{table}

We run several additional experiments to further analyze the behavior of our method. For the those experiments we use the CUB-200-2011 bird dataset, and compare to the method by Hariharan and Girshick~\cite{hariharan_low-shot_2017} in Table \ref{tab:results2}.

We first report the baseline model in the top-1, 1-shot scenario. Due to the relative shallowness of the classification network and without any sample selection or hallucination, the performance is quite low. 

Methods using simple nearest neighbour classifiers can perform well on few-shot learning tasks~\cite{krishnanmax}. We implement a simple nearest-neigbour classifier using the representations learned in our baseline on the base class samples, $x_i \in S_{train}^{base}$, specifically making use of the last hidden layer of the network. This model marginally outperforms the baseline. 

Improving the novel class data diversity by using self-paced sample selection and k-means clustering-and-dismissal, the performance rises by 5.3 points to 14.4, which equals more than 50\% relative improvement. 

So far, we have used a classifier with simple architecture and loss function in order to present the most general possible framework and to allow for a fair comparison with baseline methods. However, we expect a significant boost in accuracy using larger classifiers. To test this hypothesis, we fine-tune a modified ResNet-18~\cite{he2016deep}.
We first reduce the output dimensionality of the last pooling layer from 512 to 256 by lowering the amount of filters. After having trained this model on the base classes, we replace the last, fully-connected layer of size $|\Cbase|$ with a smaller one of size $|\Cnovel|$ to account for the different amount of classes. Afterwards, we freeze all layers except the final one, and train with $S_{train}^{novel} \cup S^{novel}_{gen}$ after having ranked the existing samples with the best shallow network. We observe comparable results to Hariharan and Girshick~\cite{hariharan_low-shot_2017} despite of neither having used the ResNet-18 as a ranking function for self-paced learning, nor performing iterative sampling. Note that our method provides a general framework to augment the training set with class-discriminative generated samples that can potentially be used in conjunction with more sophisticated methods as the SGM loss~\cite{hariharan_low-shot_2017} to obtain better results.

\section{Conclusion and Future Work}
In this paper, we proposed to extend few-shot learning by incorporating image hallucination from 3D models in conjunction with a self-paced learning strategy. Experiments on the CUB dataset demonstrate that learning generative methods employing 3D models reaches performance that significantly outperforms our baseline and is competitive to popular methods in the field. Thus the proposed approach allows for an efficient compensation of the lack of data in novel categories. 

For future work we plan to optimize the pipeline in an end-to-end fashion, discarding the self-paced learning sample selection and replacing it with learnable viewpoint angle parameters.

{\small
\bibliographystyle{ieee}
\bibliography{main}
}

\end{document}